\documentclass{article} % For LaTeX2e
\usepackage{iclr2017_conference,times}
\usepackage{hyperref}
\usepackage{url}
\usepackage{mathtools}
\usepackage{tikz}
\usepackage{subcaption}
\usepackage{circuitikz}
\usepackage{amssymb}
\usepackage{multirow}
\usepackage{amsmath}
\usepackage{color}

\usepackage{booktabs}
\usepackage{amsfonts}
\usepackage{amsmath}
\usepackage{amssymb}
\usepackage{multirow}
\usepackage{verbatim}

\title{Neural Architecture Search with\\Reinforcement Learning}
\author{Barret Zoph\thanks{Work done as a member of the Google Brain Residency program (\url{g.co/brainresidency}.)}, ~Quoc V. Le \\
Google Brain \\
\texttt{\{barretzoph,qvl\}@google.com} \\
}

\begin{document}

\maketitle

\begin{abstract}
Neural networks are powerful and flexible models that work well for many difficult learning tasks in image, speech and natural language understanding. Despite their success, neural networks are still hard to design.
%neural networks still require much domain knowledge in designing architectures. 
In this paper, we
use a recurrent network to generate the model descriptions of neural networks and train this RNN with reinforcement learning to maximize the expected accuracy of the generated architectures on a validation set. 
%formulate this task as predicting configuration strings, where each string can specify all information needed to construct a network such as connectivity between layers as well as their hyperparameters. In our framework, these configuration strings are variable-length and thus can be generated by a recurrent network. A child network is constructed given a string and trained to convergence. Using a desired performance metric of the child network as the reward signal, we can compute policy gradient to update the main recurrent network such that it will sample better architectures in the next iteration. 
%In our experiments, this formulation proves to be effective given enough computation. 
On the CIFAR-10 dataset, our method, starting from scratch, can design a novel network architecture that rivals the best human-invented architecture in terms of test set accuracy. Our CIFAR-10 model achieves a test error rate of $3.65$, which is $0.09$ percent better and 1.05x faster than the previous state-of-the-art model that used a similar architectural scheme. On the Penn Treebank dataset, our model can compose a novel recurrent cell that outperforms the widely-used LSTM cell, and other state-of-the-art baselines.  Our cell achieves a test set perplexity of 62.4 on the Penn Treebank, which is 3.6 perplexity better than the previous state-of-the-art model. The cell can also be transferred to the character language modeling task on PTB and achieves a state-of-the-art perplexity of 1.214.

%{\color{red} Now lets mention we also get SOTA on character PTB task with a perplexity of 1.214.}

%Finding good neural network architectures is essential in many practical applications, and is an open research problem. Currently, a large fraction of deep learning research is coming up with new architectures. In this paper, we try to find good neural network architectures, including its hyper parameters, automatically (starting with little prior knowledge). Prior work on finding hyper parameters for neural networks has been limited to a much smaller space, such as tuning only the learning rate, weight initialization, etc ... In this paper we greatly expand the search space, which allows for a lot less human intervention. We automatically find good convolution architectures on the Cifar10 and Cifar100 datasets that outperform existing methods hyperparameters optimization methods by a significant margin and discover interesting properties of convolutions.
\end{abstract}

\section{Introduction}
The last few years have seen much success of deep neural networks in many challenging applications, such as speech recognition~\citep{hinton2012deep}, image recognition~\citep{lecun1998gradient,krizhevsky2012imagenet} and machine translation~\citep{sutskever2014sequence,bahdanau2014neural,wu2016google}. Along with this success is a paradigm shift from feature designing to architecture designing, i.e., from SIFT~\citep{lowe1999object}, and HOG~\citep{dalal2005histograms}, to AlexNet~\citep{krizhevsky2012imagenet}, VGGNet~\citep{simonyan2014very}, GoogleNet~\citep{szegedy2015going}, and ResNet~\citep{he2015deep}. Although it has become easier, designing architectures still requires a lot of expert knowledge and takes ample time. %still takes time, and requires human experts, with a lot of experimentation.

%There are existing techniques that aim at simplifying the process of architecture search, especially hyperparameter selection. Currently, grid search, random search~\citep{bergstra2012random},  and Bayesian optimization~\citep{snoek2012practical} are standard ways to optimize hyperparameters, such as learning rates, number of hidden units, or number of layers. While these methods are good at tuning hyperparameters, it is currently not possible to ask them to compose a novel architecture from scratch. In fact, most of them still rely on a good initial model, designed by a human expert. 

%rely on many heuristics to be able to find good architectures. 

\begin{figure}[h!]
\begin{center}
\centerline{\includegraphics[width=0.6\columnwidth]{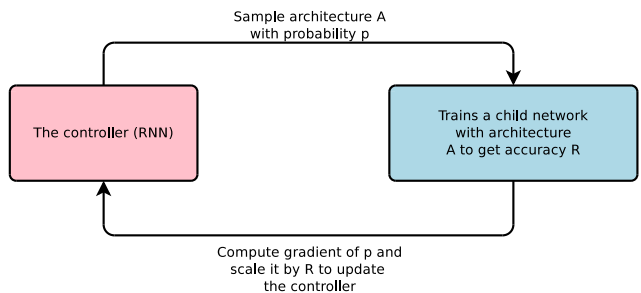}}
\caption{An overview of Neural Architecture Search.}
\label{figure:NAS}
\end{center}
\end{figure} 

This paper presents Neural Architecture Search, a gradient-based method for finding good architectures (see Figure~\ref{figure:NAS}) . Our work is based on the observation that the structure and connectivity of a neural network can be typically specified by a variable-length string. It is therefore possible to use a recurrent network -- the controller -- to generate such string. Training the network specified by the string -- the ``child network'' -- on the real data will result in an accuracy on a validation set. Using this accuracy as the reward signal, we can compute the policy gradient to update the controller. As a result, in the next iteration, the controller will give higher probabilities to architectures that receive high accuracies. In other words, the controller will learn to improve its search over time. %Figure shows an overview of our approach.

%To speed up the training, we allow the controller to sample many models at a time, and trains them in parallel. The parameters of the controller are updated asynchronously by policy gradients coming from the child models, in a similar manner to Downpour SGD~\citep{dean2012large}. 
%The fact that gradient-based techniques can be updated asynchronously in parallel is a strong advantage for our method against other search-based methods, where parallel updates are more difficult.

Our experiments show that Neural Architecture Search can design good models from scratch, an achievement considered not possible with other methods. On image recognition with CIFAR-10, Neural Architecture Search can find a novel ConvNet model that is better than most human-invented architectures. Our CIFAR-10 model achieves a 3.65 test set error, while being 1.05x faster than the current best model. On language modeling with Penn Treebank, Neural Architecture Search can design a novel recurrent cell that is also better than previous RNN and LSTM architectures. The cell that our  model found achieves a  test set perplexity of 62.4 on the Penn Treebank dataset, which is 3.6 perplexity better than the previous state-of-the-art.

%The current standard approach for 

%It is currently possible to tune hyperparameters

%Deep neural networks have enjoyed much success in the last few years in many domains: image recognition, speech recognition, 

%Deep learning has seen a lot of success in the past few years. One of the biggest unsolved problems in deep learning is coming up with new architectures for a given dataset. This is a challenging line of research where humans are frequently coming up with new architectures that outperform previous ones. Our goal in this paper is to try to automatically find good architectures, searching over a massive space.

\section{Related Work}
%\label{gen_inst}

Hyperparameter optimization is an important research topic in machine learning, and is widely used in practice~\citep{bergstra2011algorithms,bergstra2012random,snoek2012practical,snoek2015scalable, saxena2016convolutional}. Despite their success, these methods are still limited in that they only search models from a fixed-length space. In other words, it is difficult to ask them to generate a variable-length configuration that specifies the structure and connectivity of a network. In practice, these methods often work better if they are supplied with a good initial model~\citep{bergstra2012random,snoek2012practical,snoek2015scalable}. There are Bayesian optimization methods that allow to search non fixed length architectures~\citep{bergstra2013making,mendoza2016towards}, but they are less general and less flexible than the method proposed in this paper.

Modern neuro-evolution algorithms, e.g.,~\cite{wierstra2005modeling,floreano2008neuroevolution,stanley2009hypercube}, on the other hand, are much more flexible for composing novel models, yet they are usually less practical at a large scale. Their limitations lie in the fact that they are search-based methods, thus they are slow or require many heuristics to work well. 

Neural Architecture Search has some parallels to program synthesis and inductive programming, the idea of searching a program from examples~\citep{summers1977methodology,biermann1978inference}. In machine learning, probabilistic program induction has been used successfully in many settings, such as learning to solve simple Q\&A~\citep{liang2010learning,neelakantan2015neural,andreas2016learning}, sort a list of numbers~\citep{reed2015neural}, and learning with very few examples~\citep{lake2015human}. 
%Despite their distinguished history, many inductive programming methods have yet to deliver at scale, perhaps because they are slow and require many search heuristics to work well. 

The controller in Neural Architecture Search is auto-regressive, which means it predicts hyperparameters one a time, conditioned on previous predictions. This idea is borrowed from the decoder in end-to-end sequence to sequence learning~\citep{sutskever2014sequence}. Unlike sequence to sequence learning, our method optimizes a non-differentiable metric, which is the accuracy of the child network. It is therefore similar to the work on BLEU optimization in Neural Machine Translation~\citep{ranzato2015sequence,ShenCHHWSL15}. Unlike these approaches, our method learns directly from the reward signal without any supervised bootstrapping. 
%Finally, to dynamically specify connections from one layer to previous layers, we also make use of an attention mechanism, similar to those proposed in pointer~\citep{vinyals2015pointer} and set-selection connections~\citep{neelakantan2015neural}. 

Also related to our work is the idea of learning to learn or meta-learning~\citep{thrun2012learning}, a general framework of using information learned in one task to improve a future task. More closely related is the idea of using a neural network to learn the gradient descent updates for another network~\citep{andrychowicz2016learning} and the idea of using reinforcement learning to find update policies for another network~\citep{li2016learning}. 

%In this work, we use policy gradient to train the controller to predict architectures of neural models. The use of policy gradient for learning new optimization algorithms for neural models have been recently explored in~\citep{}.

%Finally, our work is inspired by~\citep{turing1950computing}, who suggested the concept of "learning machine" where an experimenter can intelligently design a set of experiments and mutations to teach and improve a child machine. The interaction between the controller and the child networks can be viewed as an implementation of this idea.

%He also hinted that evolution maybe too inefficient, and that the experimenter needs to find traces of problems to find better mutations for 

%- Bayesian Inference and Random Search \\
%- {\color{red} \textbf{QVL} Anything else?} Genetic programming?\\
%- Papers: 
%\begin{itemize}
%	\item http://papers.nips.cc/paper/4522-practical-bayesian-optimization-of-machine-learning-algorithms.pdf
%	\item http://www.jmlr.org/proceedings/papers/v37/snoek15.pdf
%	\item http://www.jmlr.org/papers/volume13/bergstra12a/bergstra12a.pdf
%	\item https://papers.nips.cc/paper/4443-algorithms-for-hyper-parameter-optimization.pdf
%\end{itemize}

%\subsection{Scalable Bayesian Optimization Using Deep Neural Networks}
%\begin{itemize}
%	\item Method scales linearly with the number of data points instead of cubicly like in normal bayesian optimization
%	\item best result is 93.63
%\end{itemize}

\section{Methods}
\label{sec:methods}
In the following section, we will first describe a simple method of using a recurrent network to generate convolutional architectures. We will show how the recurrent network can be trained with a policy gradient method to maximize the expected accuracy of the sampled architectures. We will present several improvements of our core approach such as forming skip connections to increase model complexity and using a parameter server approach to speed up training. In the last part of the section, we will focus on generating recurrent architectures, which is another key contribution of our paper.

\iffalse
\subsection{Architecture Descriptions}
As stated earlier, our key observation is that many common neural architectures can be summarized in variable-length strings, which we also call ``architecture descriptions.'' Below is an example architecture description of a simple 2-layer convolutional network:
\begin{verbatim}
    layer {
        name = layer_1
        type = conv
        connect_to = input
        filter_width = 2
        filter_height = 3
        stride_width = 4
        stride_height = 5
        num_filters = 6
    }
    layer {
        name = layer_2
        type = conv
        connect_to = layer_1
        filter_width = 7
        filter_ height = 8
        stride_width = 9
        stride_height = 10
        num_filters = 11
    }
    layer {
        name = classifier
        connect_to = layer_2
        num_class = 10
    }
\end{verbatim}
By focusing on key hyperparameters, the above description can be compressed into:
\begin{verbatim}
    2 3 4 5 6 7 8 9 10 11
\end{verbatim}
Notice that information regarding layer types and connectivity between layers are skipped entirely because we are only interested in feedforward convolutional networks without skip connections. The last layer is also skipped because we do not want to vary it. Although the new description does not have hyperparameter names such as ``filter\_width'', ``filter\_height'', they can be inferred given the order of the tokens.   In the following section, we will show how we use a recurrent network to generate this type of architecture descriptions.

\fi

\subsection{Generate Model Descriptions with a Controller Recurrent Neural Network}
\label{sec:Controller_RNN}

In Neural Architecture Search, we use a controller to generate architectural hyperparameters of neural networks. To be flexible, the controller is implemented as a recurrent neural network. Let's suppose we would like to predict feedforward neural networks with only convolutional layers, we can use the controller to generate their hyperparameters as a sequence of tokens:  
%In this case, the string follows the grammar that it should consist of first some global hyperparameters, e.g. learning rates, then a list of layer blocks where each block contains all information about that layer. 

%Therefore, for each architecture, the controller needs to generate a configuration string which specifies the layers and connectivity between layers. At each token in the string, the controller RNN will predict an aspect of the architecture, conditioned on what it has sampled so far, until the entire architecture has been predicted. 

\begin{figure}[h!]
\begin{center}
\centerline{\includegraphics[width=0.7\columnwidth]{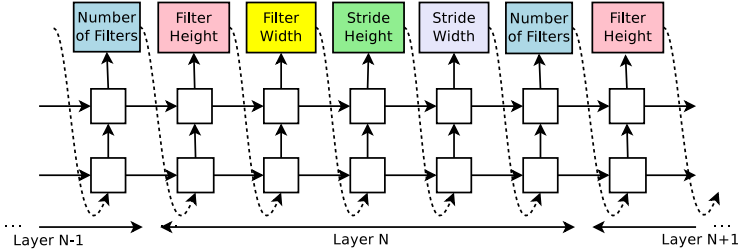}}
\caption{How our controller recurrent neural network samples a simple convolutional network. It predicts filter height, filter width, stride height, stride width, and number of filters for one layer and repeats. Every prediction is carried out by a softmax classifier and then fed into the next time step as input.}
\label{figure:Controller_RNN}
\end{center}
\end{figure}

%This generation process is auto-regressive: the controller RNN generates one token at a time conditioned on previous tokens until all needed tokens are generated. The controller RNN needs to predict hyperparameters for one layer at a time where every prediction is carried out by a softmax classifier.

%Every prediction is carried out by a softmax classifier, whose parameters are shared with other softmax classifiers that predict the same hyperparameter type. For example, the parameters for the softmax that predicts "Number of Filters" for all layers are shared, but are different from parameters for the softmax that predicts "Filter Height". Here we assume that the generated architecture has layer $N$ directly sends it output to layer $N+1$.

%Our assumption is that the generated architecture has layer $N$ directly sends it output to layer $N+1$. As this is restrictive,   we will revisit this issue in Section~\ref{sec:skip_connection}, and propose a way to have skip connections in our framework.

In our experiments, the process of generating an architecture stops if the number of layers exceeds a certain value. This value follows a schedule where we increase it as training progresses. Once the controller RNN finishes generating an architecture, a neural network  with this architecture is built and trained. At convergence, the accuracy of the network on a held-out validation set is recorded. The parameters of the controller RNN, $\theta_c$, are then optimized in order to maximize the expected validation accuracy of the proposed architectures. In the next section, we will describe a policy gradient method which we use to update parameters $\theta_c$ so that the controller RNN generates better architectures over time.

%The underlying assumption in this section is that layer $N+1$ connects to layer $N$, which is quite restrictive. In section~\ref{sec:skip_connection}, we will revisit this issue and propose a way to have skip connections in our framework.

\subsection{Training with REINFORCE}
The list of tokens that the controller predicts can be viewed as a list of actions $a_{1:T}$ to design an architecture for a child network. At convergence, this child network will achieve an accuracy $R$ on a held-out dataset. We can use this accuracy $R$ as the reward signal and use reinforcement learning to train the controller. More concretely, to find the optimal architecture, we ask our controller to maximize its expected reward, represented by $J(\theta_c)$:

$$J(\theta_c) = E_{P(a_{1:T};\theta_c)}[R] $$

Since the reward signal $R$ is non-differentiable, we need to use a policy gradient method to iteratively update $\theta_c$. In this work, we use the REINFORCE rule from~\cite{Williams92simplestatistical}: 
$$ \bigtriangledown_{\theta_c} J(\theta_c) = \sum_{t=1}^{T} E_{P(a_{1:T};\theta_c)}\big[\bigtriangledown_{\theta_c} \log P(a_t|a_{(t-1):1};\theta_c)R\big] $$

An empirical approximation of the above quantity is:
$$ \frac{1}{m} \sum_{k=1}^{m} \sum_{t=1}^{T} \bigtriangledown_{\theta_c} \log P(a_t|a_{(t-1):1};\theta_c)R_{k} $$

Where $m$ is the number of different architectures that the controller samples in one batch and $T$ is the number of hyperparameters our controller has to predict to design a neural network architecture. The validation accuracy that the $k$-th neural network architecture achieves after being trained on a training dataset is $R_k$.

The above update is an unbiased estimate for our gradient, but has a very high variance. In order to reduce the variance of this estimate we employ a baseline function:

$$ \frac{1}{m} \sum_{k=1}^{m} \sum_{t=1}^{T} \bigtriangledown_{\theta_c} \log P(a_t|a_{(t-1):1};\theta_c)(R_{k} - b) $$

As long as the baseline function $b$ does not depend on the on the current action, then this is still an unbiased gradient estimate. In this work, our baseline $b$ is an exponential moving average of the previous architecture accuracies.

%Finally, note that in this work, we only consider accuracy as the desired performance metric, but it is possible to generalize the framework to add training time as another factor~\citep{snoek2012practical}. For example, we can consider $\frac{R}{time\_to\_convergence}$ as the performance metric, which encourages architectures that are fast to train and achieve good accuracy. 

%\subsection{Accelerate Training with Parallelism and Asynchronous Updates}

\begin{figure}[h!]
\begin{center}
\centerline{\includegraphics[width=0.8\columnwidth]{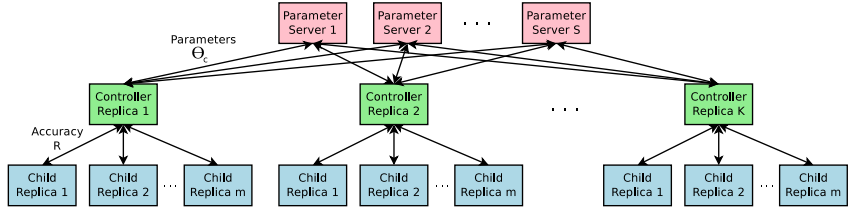}}
\caption{Distributed training for Neural Architecture Search. We use a set of $S$ parameter servers to store and send parameters to $K$ controller replicas. Each controller replica then samples $m$ architectures and run the multiple child models in parallel. The accuracy of each child model is recorded to compute the gradients with respect to $\theta_c$, which are then sent back to the parameter servers.}
\label{figure:Dist_Setup}
\end{center}
\end{figure}

\paragraph{Accelerate Training with Parallelism and Asynchronous Updates:} In Neural Architecture Search, each gradient update to the controller parameters $\theta_c$ corresponds to training one child network to convergence. As training a child network can take hours, we use distributed training and asynchronous parameter updates in order to speed up the learning process of the controller~\citep{dean2012large}. We use a parameter-server scheme where we have a parameter server of $S$ shards, that store the shared parameters for $K$ controller replicas. Each controller replica samples $m$ different child architectures that are trained in parallel. The controller then collects gradients according to the results of that minibatch of $m$ architectures at convergence and sends them to the parameter server in order to update the weights across all controller replicas. In our implementation, convergence of each child network is reached when its training exceeds a certain number of epochs. This scheme of parallelism is summarized in Figure~\ref{figure:Dist_Setup}.

\subsection{Increase Architecture Complexity with Skip Connections and Other Layer Types}
\label{sec:skip_connection}

In Section~\ref{sec:Controller_RNN}, 
%we assume that layer $N+1$ directly connects with layer $N$. This assumption is restrictive because it constrains 
the search space does not have skip connections, or branching layers used in modern architectures such as GoogleNet~\citep{szegedy2015going}, and Residual Net~\citep{he2015deep}. In this section we introduce a method that allows our controller to propose skip connections or branching layers, thereby widening the search space. 

\begin{figure}[h!]
\begin{center}
\centerline{\includegraphics[width=0.75\columnwidth]{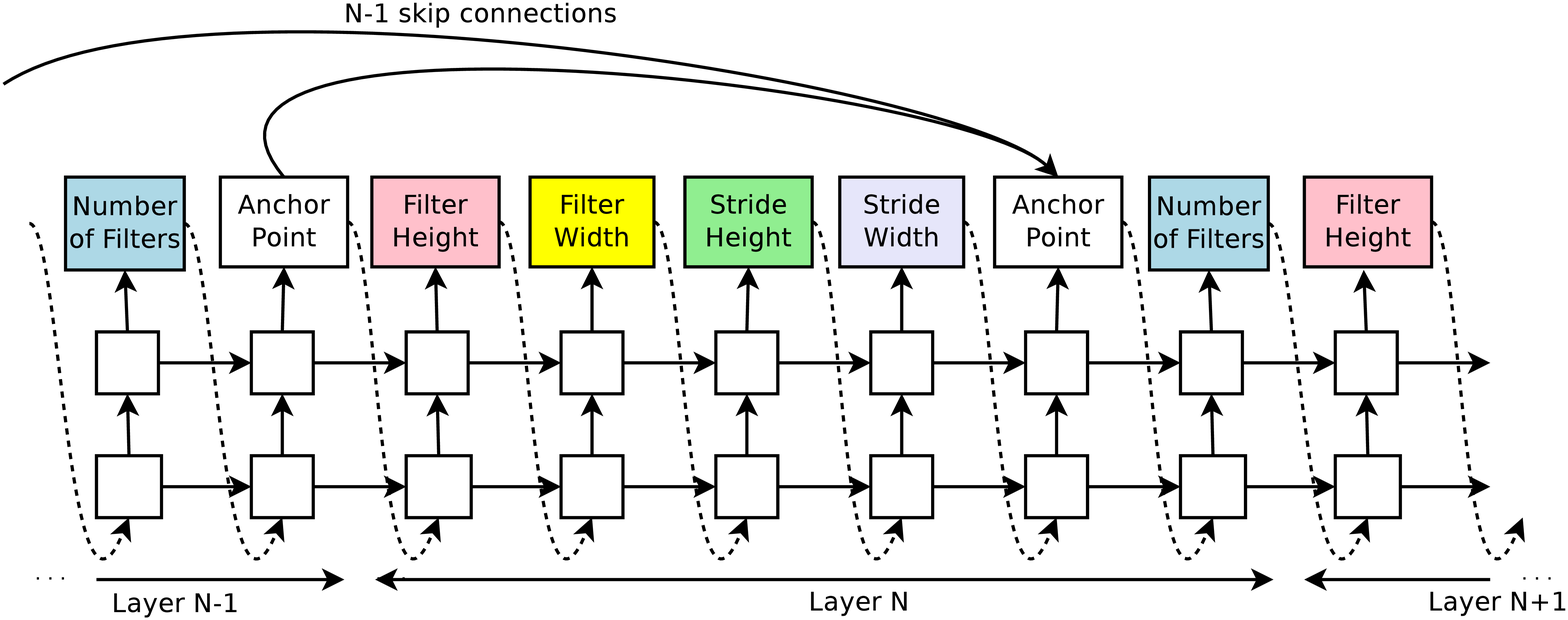}}
\caption{The controller uses anchor points, and set-selection attention to form skip connections.}
\label{figure:Controller_RNN_PTR}
\end{center}
\end{figure}

To enable the controller to predict such connections, we use a set-selection type attention~\citep{neelakantan2015neural} which was built upon the attention mechanism~\citep{bahdanau2014neural,vinyals2015pointer}. At layer $N$, we add an anchor point which has $N-1$ content-based sigmoids to indicate the previous layers that need to be connected. Each sigmoid is a function of the current hiddenstate of the controller and the previous hiddenstates of the previous $N-1$ anchor points:

$$\mathrm{P(Layer \ j \ is \ an \ input \ to \ layer \ i) = \ } \mathrm{sigmoid}(v^\mathrm{T} \mathrm{tanh}(W_{prev}*h_{j} + W_{curr} * h_{i})), $$
 where $h_j$ represents the hiddenstate of the controller at anchor point for the $j$-th layer, where $j$ ranges from $0$ to $N-1$. We then sample from these sigmoids to decide what previous layers to be used as inputs to the current layer. The matrices $W_{prev}$,  $W_{curr}$ and $v$ are trainable parameters. As these connections are also defined by probability distributions, the REINFORCE method still applies without any significant modifications. 
Figure~\ref{figure:Controller_RNN_PTR} shows how the controller uses skip connections to decide what layers it wants as inputs to the current layer.

In our framework, if one layer has many input layers then all input layers are concatenated in the depth dimension. Skip connections can cause ``compilation failures'' where one layer is not compatible with another layer, or one layer may not have any input or output. To circumvent these issues, we employ three simple techniques. First, if a layer is not connected to any input layer then the image is used as the input layer. Second, at the final layer we take all layer outputs that have not been connected and concatenate them before sending this final hiddenstate to the classifier. Lastly, if input layers to be concatenated have different sizes, we pad the small layers with zeros so that the concatenated layers have the same sizes. 

Finally, in Section~\ref{sec:Controller_RNN}, we do not predict the learning rate and we also assume that the architectures consist of only convolutional layers, which is also quite restrictive. It is possible to add the learning rate as one of the predictions. Additionally, it is also possible to predict pooling, local contrast normalization~\citep{jarrett2009best,krizhevsky2012imagenet}, and batchnorm~\citep{BatchNorm} in the architectures. To be able to add more types of layers, we need to add an additional step in the controller RNN to predict the layer type, then other hyperparameters associated with it. 
%\subsection{Curriculum Learning}
%\label{sec:curriculum}
%{\color{red} We are doing 1600 models at each layer and then fix the number of layers it samples to be 13 for the max pooling and 20 for the stride. We never predict a variable number of layers in these experiments.}
%In our experiments, if we allow the controller RNN to predict when it wants to stop, it tends to prefer sampling shallow architectures, as deeper networks can result in more failures, slow training or low accuracies. We circumvent this issue by asking the controller RNN to sample architectures with an increasing number of layers as training progresses. On CIFAR-10, the curriculum is to increase the depth by 2 layers after every 1600 samples starting at depth 6 and ending at depth 20.  We find that curriculum learning significantly improves the performance of our controller RNN. 

\subsection{Generate Recurrent Cell Architectures}
\label{sec:recurrent_cell}
In this section, we will modify the above method to generate recurrent cells. At every time step $t$, the controller needs to find a functional form for $h_t$ that takes $x_t$ and $h_{t-1}$ as inputs. The simplest way is to have $h_t = \tanh(W_1 * x_t + W_2 * h_{t-1})$, which is the formulation of a basic recurrent cell. A more complicated formulation is the widely-used LSTM recurrent cell~\citep{lstm}.

The computations for basic RNN and LSTM cells can be generalized as a tree of steps that take $x_t$ and $h_{t-1}$ as inputs and produce $h_t$ as final output. The controller RNN needs to label each node in the tree with a combination method (addition, elementwise multiplication, etc.) and an activation function ($\tanh$, $\mathrm{sigmoid}$, etc.) to merge two inputs and produce one output. Two outputs are then fed as inputs to the next node in the tree. To allow the controller RNN to select these methods and functions, we index the nodes in the tree in an order so that the controller RNN can visit each node one by one and label the needed hyperparameters.

Inspired by the construction of the LSTM cell~\citep{lstm}, we also need cell variables $c_{t-1}$ and $c_t$ to represent the memory states. To incorporate these variables, we need the controller RNN to predict what nodes in the tree to connect these two variables to. These predictions can be done in the last two blocks of the controller RNN. 

\begin{figure}[h!]
\begin{center}
\includegraphics[width=0.21\columnwidth]{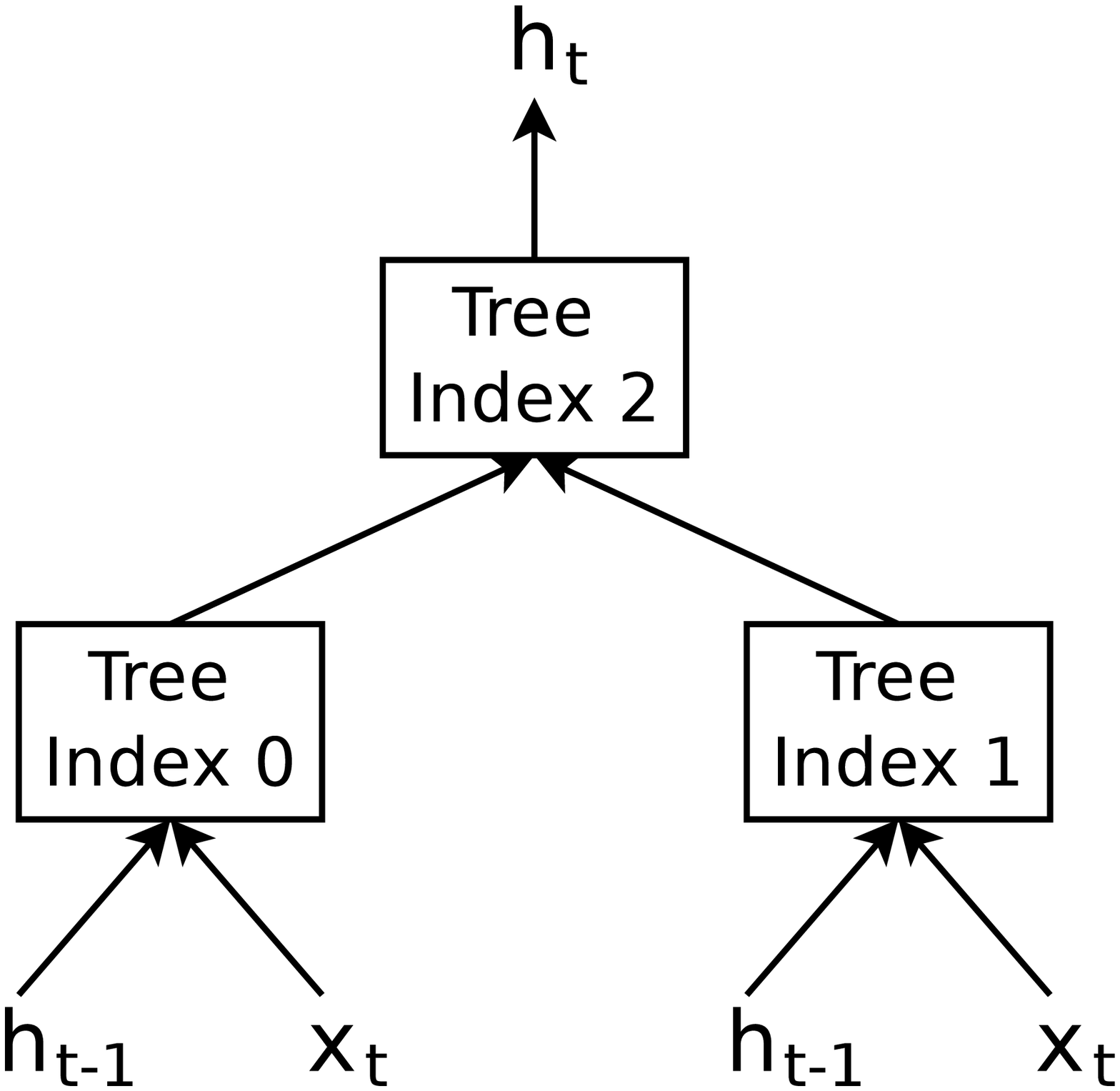}
\includegraphics[width=0.53\columnwidth]{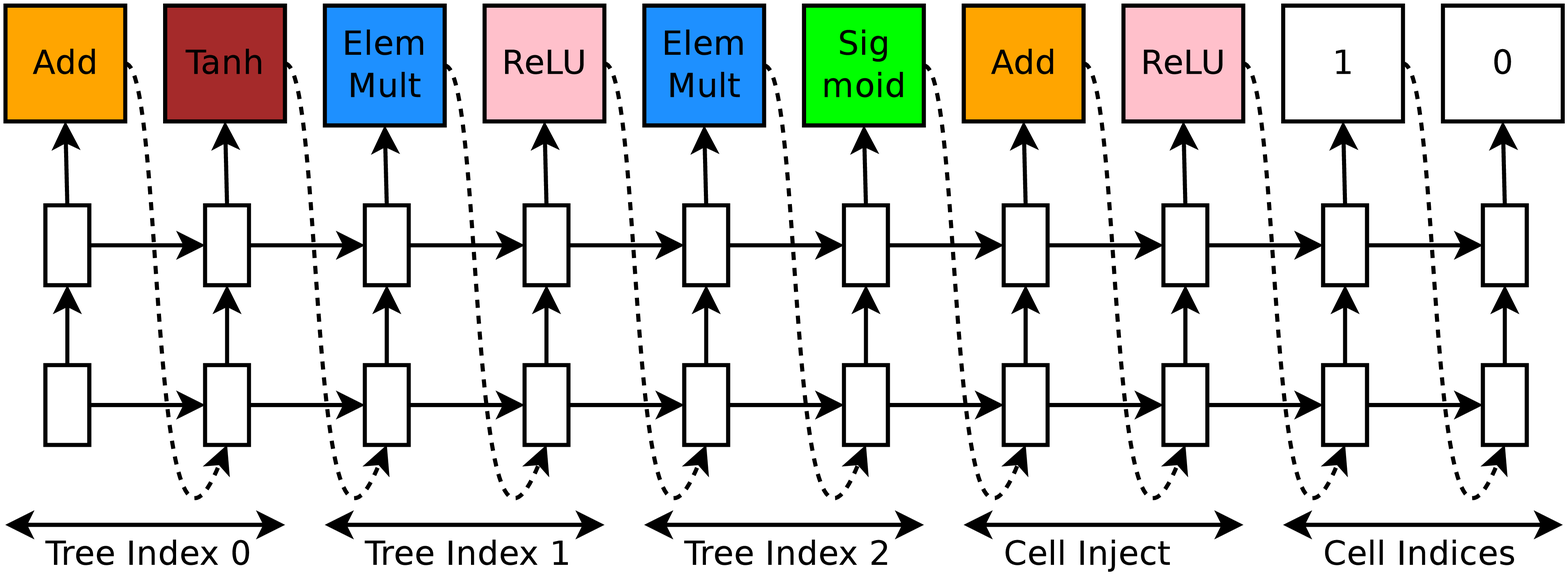}
\includegraphics[width=0.24\columnwidth]{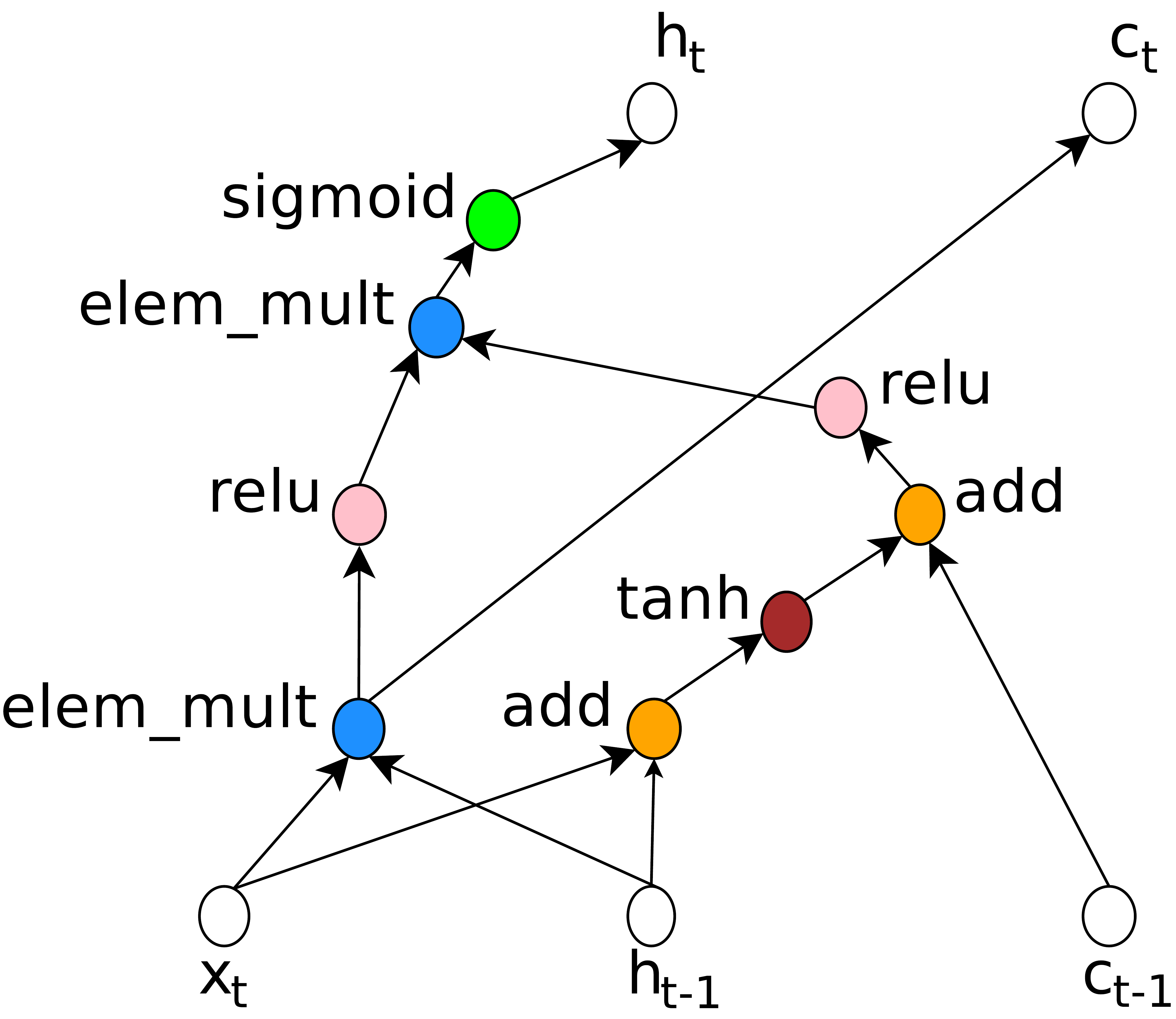}
\caption{An example of a recurrent cell constructed from a tree that has two leaf nodes (base 2) and one internal node. Left: the tree that defines the computation steps to be predicted by controller. Center: an example set of predictions made by the controller for each computation step in the tree.  Right: the computation graph of the recurrent cell constructed from example predictions of the controller. }
\label{figure:RNN_Method}
\end{center}
\end{figure} 
To make this process more clear, we show an example in Figure~\ref{figure:RNN_Method}, for a tree structure that has two leaf nodes and one internal node. The leaf nodes are indexed by 0 and 1, and the internal node is indexed by 2. The controller RNN needs to first predict 3 blocks, each block specifying a combination method and an activation function for each tree index. After that it needs to predict the last 2 blocks that specify how to connect $c_t$ and $c_{t-1}$ to temporary variables inside the tree. Specifically, according to the predictions of the controller RNN in this example, the following computation steps will occur:
\begin{itemize}
\item The controller predicts $Add$ and $Tanh$ for tree index 0, this means we need to compute $a_0 = \tanh(W_1 * x_t + W_2 * h_{t-1})$. 
\item The controller predicts $Elem Mult$ and $ReLU$ for tree index 1, this means we need to compute $a_1 = \mathrm{ReLU}\big((W_3 * x_t) \odot (W_4 * h_{t-1})\big)$. 
\item The controller predicts 0 for the second element of the ``Cell Index'', $Add$ and $ReLU$ for elements in ``Cell Inject'', which means we need to compute $a_0^{new} = \mathrm{ReLU}(a_0 + c_{t-1})$. Notice that we don't have any learnable parameters for the internal nodes of the tree.
\item The controller predicts $Elem Mult$ and $Sigmoid$ for tree index 2, this means we need to compute $a_2 = \mathrm{sigmoid}( a_0^{new} \odot a_1)$. Since the maximum index in the tree is 2, $h_t$ is set to $a_2$.
\item The controller RNN predicts 1 for the first element of the ``Cell Index'', this means that we should set $c_t$ to the output of the tree at index 1 before the activation, i.e., $c_t = (W_3 * x_t) \odot (W_4 * h_{t-1})$.
\end{itemize}

In the above example, the tree has two leaf nodes, thus it is called a ``base 2'' architecture. In our experiments, we use a base number of 8 to make sure that the cell is expressive. 

%We also find that learning recurrent cells is slightly more stable than learning convolutional architectures. For that reason, we do not use a schedule during training in our cell learning experiments.

%Note that the computation cost of the cell does not depend on the Base number if the number of parameters is fixed. This is because the dominant computation cost of the cells is associated with matrix multiplications, which occur at the leaf nodes, and only depend on the number of parameters. 

%s of the tree, where matrix multiplications are done and parameters are kept.  

%In our experiments, we keep our results comparable with prior work by holding the number of parameters fixed.

%Therefore, to keep the results fair, as we increase the Base number, we keep the number of parameters fixed. This way, the computational complexity of the cells do not vary.

\section{Experiments and Results}
\label{others}

We apply our method to an image classification task with CIFAR-10 and a language modeling task with Penn Treebank, two of the most benchmarked datasets in deep learning. On CIFAR-10, our goal is to find a good convolutional architecture whereas on Penn Treebank our goal is to find a good recurrent cell. On each dataset, we have a separate held-out validation dataset to compute the reward signal. The reported performance on the test set is computed only once for the network that achieves the  best result on the held-out validation dataset. More details about our experimental procedures and results are as follows.

%On the CIFAR-10 dataset, our method, starting from scratch, finds a new convolutional architecture that performs near the level of best human-invented models, which took many years to develop. On the Penn Treebank dataset, our method, also starting from scratch, invents a new recurrent cell that beats the LSTM cell, and other state-of-the-art models. 

\subsection{Learning Convolutional Architectures for CIFAR-10}

\paragraph{Dataset:} In these experiments we use the CIFAR-10 dataset with data preprocessing and augmentation procedures that are in line with other previous results. We first preprocess the data by whitening all the images. Additionally, we upsample each image then choose a random 32x32 crop of this upsampled image. Finally, we use random horizontal flips on this 32x32 cropped image.

\paragraph{Search space:} Our search space consists of convolutional architectures, with rectified linear units as non-linearities~\citep{nair2010rectified}, batch normalization~\citep{BatchNorm} and skip connections between layers (Section~\ref{sec:skip_connection}). For every convolutional layer, the controller RNN has to select a filter height in [1, 3, 5, 7], a filter width in [1, 3, 5, 7], and a number of filters in [24, 36, 48, 64].
%\begin{itemize}
%    \item Filter Height: [1, 3, 5, 7]
%    \item Filter Width: [1, 3, 5, 7]
%    \item Number of Filters: [24, 36, 48, 64] %{\color{red} the filters used for the max pooling were [6, 12, 24, 36]}
%\end{itemize}
For strides, we perform two sets of experiments, one where we fix the strides to be 1, and one where we allow the controller to predict the strides in [1, 2, 3]. 
%Notice that for CIFAR-10, as the images are small, fixing the strides at 1 is a good option to limit the search space; whereas predicting strides require us to apply heuristics such as zero padding and depth concatenation to the child models, so the controller will have a difficult time to find a good architecture. Comparing the results of the two sets of experiments can help us shed light on how well controller performs when little knowledge of the dataset is present.

%\subsubsection{Training Details}

\paragraph{Training details:} The controller RNN is a two-layer LSTM with 35 hidden units on each layer. It is trained with the ADAM optimizer~\citep{ADAM} with a learning rate of 0.0006. The weights of the controller are initialized uniformly between -0.08 and 0.08. For the distributed training, we set the number of parameter server shards $S$ to 20, the number of controller replicas $K$ to 100 and the number of child replicas $m$ to 8, which means there are 800 networks being trained on 800 GPUs concurrently at any time.

Once the controller RNN samples an architecture, a child model is constructed and trained for 50 epochs. The reward used for updating the controller is the maximum validation accuracy of the last 5 epochs cubed. The validation set has 5,000 examples randomly sampled from the training set, the remaining 45,000 examples are used for training. The settings for training the CIFAR-10 child models are the same with those used in~\cite{Huang2016Densely}. We use the Momentum Optimizer with a learning rate of 0.1, weight decay of 1e-4, momentum of 0.9 and used Nesterov Momentum ~\citep{icml2013_sutskever13}. %{\color{red} How the validation set is selected, what's the size?}

% Towards the end of its training, the controller has very a peaked distribution that forces it to keep sampling the same architectures many times. To alleviate this problem, one in every 10 batches we add noise to the sampled architecture: for each element of the architecture we randomly increment/decrement its value with probability 0.3. 

%As stated earlier, during training, the controller RNN prefers sampling architectures that have a small number of layers, as shallower networks have a higher chance of success. To alleviate this problem, 

During the training of the controller, we use a schedule of increasing number of layers in the child networks as training progresses. On CIFAR-10, we ask the controller to increase the depth by 2 for the child models every 1,600 samples, starting at 6 layers. % and ending at 20 layers.

\paragraph{Results:} After the controller trains 12,800 architectures, we find the architecture that achieves the best validation accuracy. We then run a small grid search over learning rate, weight decay, batchnorm epsilon and what epoch to decay the learning rate. The best model from this grid search is then run until convergence and we then compute the test accuracy of such model and summarize the results in Table~\ref{tab:cifar10}. As can be seen from the table, Neural Architecture Search can design several promising architectures that perform as well as some of the best models on this dataset.
\begin{table}[h!]
\centering
\small
%\begin{small}
\resizebox{\textwidth}{!}{%
\begin{tabular}{l|cc|c}
\toprule
\multicolumn{1}{c|}{\bf Model} & \multicolumn{1}{l}{\bf Depth} & \multicolumn{1}{l|}{\bf Parameters}  & \bf Error rate (\%)  \\ \midrule
Network in Network \citep{netinnet} & - & - & 8.81   \\
All-CNN \citep{allcnn} & - & - & 7.25 \\
Deeply Supervised Net \citep{dsn} & - & - & 7.97   \\
Highway Network \citep{highway} & - & -  & 7.72  \\ 
Scalable Bayesian Optimization \citep{snoek2015scalable} & - & - & 6.37 \\ \midrule
FractalNet \citep{fractalnet} & 21 & 38.6M  & 5.22  \\
with Dropout/Drop-path & 21 & 38.6M  & 4.60   \\ \midrule
ResNet \citep{he2015deep} & 110 & 1.7M  & 6.61   \\ \midrule
ResNet (reported by \citet{stochastic}) & 110 & 1.7M  & 6.41  \\ \midrule
ResNet with Stochastic Depth \citep{stochastic} & 110 & 1.7M  & 5.23   \\
 & 1202 & 10.2M & 4.91  \\ \midrule
Wide ResNet \citep{wide} & 16 & 11.0M &  4.81  \\
 & 28 & 36.5M & 4.17  \\
\midrule
ResNet (pre-activation) \citep{identity-mappings} & 164 & 1.7M & 5.46  \\
\multicolumn{1}{c|}{} & 1001 & 10.2M & 4.62  \\ \hline
DenseNet $(L=40, k=12)$ \citet{Huang2016Densely} & 40 & 1.0M &  5.24  \\
DenseNet$(L=100, k=12)$ \citet{Huang2016Densely} & 100 & 7.0M &  4.10  \\
DenseNet $(L=100, k=24)$ \citet{Huang2016Densely} & 100 & 27.2M &  3.74  \\ 
DenseNet-BC $(L=100, k=40)$ \citet{Huang2016Densely2} & 190 & 25.6M &  3.46  \\ 
\midrule
%Random Search v1 & - & - & -  \\
Neural Architecture Search v1 no stride or pooling & 15 & 4.2M & 5.50   \\
%Controller v1 Fully Connected & 15 & 6.6M & 5.56   \\
%Controller v1 No skip connections & 15 & ???M & 7.97   \\
%Controller v1 Basic Skip connections & 15 & ???M & 7.05   \\
Neural Architecture Search v2 predicting strides & 20 & 2.5M & 6.01  \\
Neural Architecture Search v3 max pooling & 39 & 7.1M & 4.47  \\
%Neural Architecture Search v3 max pooling + more filters & 39 & 17.4M & 4.27  \\
%Neural Architecture Search v3 max pooling + more filters & 39 & 32.0M & 3.84  \\
Neural Architecture Search v3 max pooling + more filters & 39 & 37.4M & 3.65 \\
%Neural Architecture Search v3 max pooling +++ more filters & 39 & 63.6M & {\color{red} ???} \\
\bottomrule
\end{tabular}%
}
\caption{Performance of Neural Architecture Search and other state-of-the-art models on CIFAR-10.}
\label{tab:cifar10}
%\end{small}
\end{table}

%%%%%qvl

First, if we ask the controller to not predict stride or pooling, it can design a 15-layer architecture that achieves 5.50\% error rate on the test set. This architecture has a good balance between accuracy and depth. In fact, it is the shallowest and perhaps the most inexpensive architecture among the top performing networks in this table. This architecture is shown in Appendix~\ref{sec:appendix}, Figure~\ref{figure:strange_net}. A notable feature of this architecture is that it has many rectangular filters and it prefers larger filters at the top layers. Like residual networks~\citep{he2015deep}, the architecture also has many one-step skip connections. This architecture is a local optimum in the sense that if we perturb it, its performance becomes worse. For example, if we densely connect all layers with skip connections, its performance becomes slightly worse: 5.56\%. If we remove all skip connections, its performance drops to 7.97\%.

In the second set of experiments, we ask the controller to predict strides in addition to other hyperparameters. As stated earlier, this is more challenging because the search space is larger. In this case, it finds a 20-layer architecture that achieves 6.01\% error rate on the test set, which is not much worse than the first set of experiments. 
%Even though this is a worse result than the previous model, it's an important achievement because the model has to search in a much larger space. %If we take into the account that fact that the controller has to search in a larger space, this result is an important achievement.

Finally, if we allow the controller to include 2 pooling layers at layer 13 and layer 24 of the architectures, the controller can design a 39-layer network that achieves 4.47\% which is very close to the best human-invented architecture that achieves 3.74\%. To limit the search space complexity we have our model predict 13 layers where each layer prediction is a fully connected block of 3 layers. Additionally, we change the number of filters our model can predict from [24, 36, 48, 64] to [6, 12, 24, 36]. Our result can be improved to 3.65\% by adding 40 more filters  to each layer of our architecture. Additionally this model with 40 filters added is 1.05x as fast as the DenseNet model that achieves 3.74\%, while having better performance. The DenseNet model that achieves 3.46\% error rate~\citep{Huang2016Densely2} uses 1x1 convolutions to reduce its total number of parameters, which we did not do, so it is not an exact comparison.

% Our architectures are better than the best reported result of 6.37\% using Bayesian optimization~\citep{snoek2015scalable}, which had a better starting point from a good human-invented architecture. 

\subsection{Learning Recurrent Cells for Penn Treebank}

%\begin{figure}[h!]
%\begin{center}
%\centerline{\includegraphics[scale=0.5]{PTB_Reward_Plot.eps}}
%\caption{Reward over time for our replica training.}
%\label{figure:Controller_RNN_PTR}
%\end{center}
%\end{figure} 

%\begin{figure}[h!]
%\begin{center}
%\centerline{\includegraphics[width=1.0\columnwidth]{Controller_Cell.pdf}}
%\caption{Illustration of how our controller RNN predicts the RNN cell configuration with a Base of 4. The Controller RNN will predict sequentially the combination method %for index 1, non-linearity for index 1, combination for index 2, ... non-linearity for index 7}
%\label{figure:Controller_Cell}
%\end{center}
%\end{figure} 

\paragraph{Dataset:} We apply Neural Architecture Search to the Penn Treebank dataset, a well-known benchmark for language modeling. On this task, LSTM architectures tend to excel~\citep{ZarembaReg, Gal2015}, and improving them is  difficult~\citep{jozefowicz2015empirical}. As PTB is a small dataset, regularization methods are needed to avoid overfitting. First, we make use of the embedding dropout and recurrent dropout techniques proposed in~\cite{ZarembaReg} and~\citep{Gal2015}. We also try to combine them with the method of sharing Input and Output embeddings, e.g., \cite{bengio2003neural,mnih2007three}, especially \cite{SocherEmbedding} and \cite{shareEmbedding}. Results with this method are marked with ``shared embeddings.''

%The current best result on this dataset requires pointer sentinel techniques~\citep{Socher2016}, which are not used by our method. We however will make a comparison with their results because it can give us a better understanding of how such human-invented techniques perform against a machine-invented cell that does not use them.  

\paragraph{Search space:} Following Section~\ref{sec:recurrent_cell}, our controller sequentially predicts a combination method then an activation function for each node in the tree. 
For each node in the tree, the controller RNN needs to select a combination method in $[add, elem\_mult]$ and an activation method in $[identity, tanh, sigmoid, relu]$. % the following hyperparameters in a set of specified values:
%\begin{itemize}
%    \item Combination method: $[add, elem\_mult]$ %with or without $max$ 
%    \item Activation function: $[identity, tanh, sigmoid, relu]$ %with or without $sin$
%\end{itemize}
%For clarity, we will indicate the results with "max/sin" for the experiments that require the $max$ function for combination and the $sin$ function for activation.
The number of input pairs to the RNN cell is called the ``base number'' and set to 8 in our experiments. When the base number is 8, the search space is has approximately $6 \times 10^{16}$ architectures, which is much larger than 15,000, the number of architectures that we allow our controller to evaluate.

\paragraph{Training details:}
The controller and its training are almost identical to the CIFAR-10 experiments except for a few modifications: 1) the learning rate for the controller RNN is 0.0005, slightly smaller than that of the controller RNN in CIFAR-10, 2) in the distributed training, we set $S$ to 20, $K$ to 400 and $m$ to 1, which means there are 400 networks being trained on 400 CPUs concurrently at any time, 3) during asynchronous training we only do parameter updates to the parameter-server once 10 gradients from replicas have been accumulated.

In our experiments, every child model is constructed and trained for 35 epochs. Every child model has two layers, with the number of hidden units adjusted so that total number of learnable parameters approximately match the ``medium'' baselines~\citep{ZarembaReg, Gal2015}. In these experiments we only have the controller predict the RNN cell structure and fix all other hyperparameters. The reward function is  $\frac{c}{\textrm{(validation perplexity)}^2}$ where $c$ is a constant, usually set at 80.

After the controller RNN is done training, we take the best RNN cell according to the lowest validation perplexity and then run a grid search over learning rate, weight initialization, dropout rates and decay epoch. The best cell found was then run with three different configurations and sizes to increase its capacity.

\paragraph{Results:} In Table~\ref{table:PTBwordresults}, we provide a comprehensive list of architectures and their performance on the PTB dataset. As can be seen from the table, the models found by Neural Architecture Search outperform other state-of-the-art models on this dataset, and one of our best models achieves a gain of almost 3.6 perplexity. Not only is our cell is better, the model that achieves 64 perplexity is also more than two times faster because the previous best network requires running a cell 10 times per time step~\citep{Zilly2016}.

%A model with a base of 8, 20M parameters, is as good as the model with the same size in~\citet{Socher2016} while not using any of their latest pointer sentinel tricks. 

\begin{table*}[h!]
\center
\begin{small}
\begin{tabular}{l|ccc}
\toprule
\bf Model & \bf Parameters &  \bf Test Perplexity\\
\midrule
\citet{Mikolov2012} - KN-5 & 2M$^\ddagger$ & $141.2$ \\
\citet{Mikolov2012} - KN5 + cache & 2M$^\ddagger$ & $125.7$ \\
\citet{Mikolov2012} - RNN & 6M$^\ddagger$ & $124.7$ \\
\citet{Mikolov2012} - RNN-LDA & 7M$^\ddagger$ & $113.7$ \\
\citet{Mikolov2012} - RNN-LDA + KN-5 + cache & 9M$^\ddagger$  & $92.0$ \\
\citet{Pascanu2013a} - Deep RNN & 6M & $107.5$ \\
\citet{Cheng2014} - Sum-Prod Net & 5M$^\ddagger$ & $100.0$ \\
\citet{ZarembaReg} - LSTM (medium) & 20M & $82.7$ \\
\citet{ZarembaReg} - LSTM (large) & 66M & $78.4$ \\
\citet{Gal2015} - Variational LSTM (medium, untied) & 20M & $79.7$ \\
\citet{Gal2015} - Variational LSTM (medium, untied, MC) & 20M & $78.6$ \\
\citet{Gal2015} - Variational LSTM (large, untied) & 66M  & $75.2$ \\
\citet{Gal2015} - Variational LSTM (large, untied, MC) & 66M & $73.4$ \\
\citet{Kim2016} - CharCNN & 19M & $78.9$ \\
\citet{shareEmbedding} - Variational LSTM, shared embeddings& 51M & $73.2$ \\
%\citet{Zilly2016} - Variational RHN NEW (241 epochs) & 32M & $68.5$ \\
%\citet{Zilly2016} - Variational RHN NEW S(400 epochs) & 21M & $66.4$ \\
%\citet{Zilly2016} - Variational RHN NEW S MC(400 epochs) & 21M & $64.4$ \\
\citet{Socher2016} - Zoneout + Variational LSTM (medium) & 20M & $80.6$ \\
\citet{Socher2016} - Pointer Sentinel-LSTM (medium) & 21M & $70.9$ \\
\citet{SocherEmbedding} - VD-LSTM + REAL (large) & 51M & $68.5$ \\
\citet{Zilly2016} - Variational RHN, shared embeddings & 24M & $66.0$ \\
\midrule
Neural Architecture Search with base 8 & 32M & $67.9$ \\
Neural Architecture Search with base 8 and shared embeddings & 25M & $64.0$ \\
Neural Architecture Search with base 8 and shared embeddings & 54M & $62.4$ \\
\bottomrule
\end{tabular}
\caption{Single model perplexity on the test set of the Penn Treebank language modeling task. Parameter numbers with $^\ddagger$ are estimates with reference to \citet{Socher2016}.}
\label{table:PTBwordresults}
\end{small}
\end{table*}

%As our method uses the dropout methods in~\cite{Gal2015}, the most direct comparison would be against their reported results. This comparison  is shown in Table~\ref{table:direct_comparison_PTB}, which reveals that our new cell improves the LSTM cell by 6.9 perplexity in an equivalent experimental setting. This gain is achieved without any significant change in the computation complexity despite the complex structure of the new cells. This is because we keep the number of parameters between the two methods the same.

%In addition to comparing against~\citep{Gal2015}, 

The newly discovered cell is visualized in Figure~\ref{fig:strange_cell} in Appendix~\ref{sec:appendix}. The visualization reveals that the new cell has many similarities to the LSTM cell in the first few steps, such as it likes to compute $W_1 * h_{t-1} + W_2 * x_t$ several times and send them to different  components in the cell.

\paragraph{Transfer Learning Results:}
To understand whether the cell can generalize to a different task, we apply it to the character language modeling task on the same dataset. We use an experimental setup that is similar to~\cite{ha2016hypernetworks}, but use variational dropout by \cite{Gal2015}. We also train our own LSTM with our setup to get a fair LSTM baseline. Models are trained for 80K steps and the best test set perplexity is taken according to the step where validation set perplexity is the best. The results on the test set of our method and state-of-art methods are reported in Table~\ref{table:charPTB}. The results on small settings with 5-6M parameters confirm that the new cell does indeed generalize, and is better than the LSTM cell. 

\begin{table*}[h!]
\center
\begin{small}
\begin{tabular}{l|cc}
\toprule
\bf RNN Cell Type & \bf Parameters &  \bf Test Bits Per Character \\
\midrule
\citet{ha2016hypernetworks} - Layer Norm HyperLSTM & 4.92M & 1.250 \\
\citet{ha2016hypernetworks} - Layer Norm HyperLSTM Large Embeddings & 5.06M & 1.233 \\
\citet{ha2016hypernetworks} - 2-Layer Norm HyperLSTM & 14.41M & 1.219 \\
\midrule
Two layer LSTM & 6.57M & 1.243 \\
Two Layer with New Cell & 6.57M & 1.228 \\
Two Layer with New Cell & 16.28M & 1.214\\
\bottomrule
\end{tabular}
\caption{Comparison between our cell and state-of-art methods on PTB character modeling. The new cell was found on word level language modeling.  }
%{\color{red} Mention that the 16.28M parameter model was run for more steps.}}
\label{table:charPTB}
\end{small}
\end{table*}

Additionally, we carry out a larger experiment where the model has 16.28M parameters. This model has a weight decay rate of $1e-4$, was trained for 600K steps (longer than the above models) and the test perplexity is taken where the validation set perplexity is highest. We use dropout rates of 0.2 and 0.5  as described in \cite{Gal2015}, but do not use embedding dropout. We use the ADAM optimizer with a learning rate of 0.001 and an input embedding size of 128. Our model had two layers with 800 hidden units. We used a minibatch size of 32 and BPTT length of 100. With this setting, our model achieves 1.214 perplexity, which is the new state-of-the-art result on this task.

Finally, we also drop our cell into the GNMT framework~\citep{wu2016google}, which was previously tuned for LSTM cells, and train an WMT14 English $\rightarrow$ German translation model. The GNMT network has 8 layers in the encoder, 8 layers in the decoder. The first layer of the encoder has bidirectional connections. The attention module is a neural network with 1 hidden layer. When a LSTM cell is used, the number of hidden units in each layer is 1024. The model is trained in a distributed setting with a parameter sever and 12 workers. Additionally, each worker uses 8 GPUs and a minibatch of 128. We use Adam with a learning rate of 0.0002 in the first 60K training steps, and SGD with a learning rate of 0.5 until 400K steps. After that the learning rate is annealed by dividing by 2 after every 100K steps until it reaches 0.1. Training is stopped at 800K steps. More details can be found in~\cite{wu2016google}.

In our experiment with the new cell, we make no change to the above settings except for dropping in the new cell and adjusting the hyperparameters so that the new model should have the same computational complexity with the base model. The result shows that our cell, with the same computational complexity, achieves an improvement of 0.5 test set BLEU than the default LSTM cell. Though this improvement is not huge, the fact that the new cell can be used without any tuning on the existing GNMT framework is encouraging. We expect further tuning can help our cell perform better.

%{\color{red} We then ran a configuration where the model had many more parameters. The hyperparameters for this model were using a weight decay rate of $1e-4$, ran for 600K steps and the test perplexity was taken where the validation set perplexity was the highest. We use 0.2 and 0.5 dropout probabilities as described in \cite{Gal2015}, but do not use embedding dropout. We use the ADAM optimizer with a learning rate of 0.001 and an input embedding size of 128. Our model had two layers with 800 hidden units. We used a minibatch size of 32 and BPTT length of 100. This was run longer (600K steps) than the LSTM comparison.}

\paragraph{Control Experiment 1 -- Adding more functions in the search space:} To test the robustness of Neural Architecture Search, we add $max$ to the list of combination functions and $sin$ to the list of activation functions and rerun our experiments. The results show that even with a bigger search space, the model can achieve somewhat comparable performance. The best architecture with $max$ and $sin$ is shown in Figure~\ref{fig:strange_cell} in Appendix~\ref{sec:appendix}.

\begin{figure}[h!]
\begin{center}
  \includegraphics[width=0.48\textwidth]{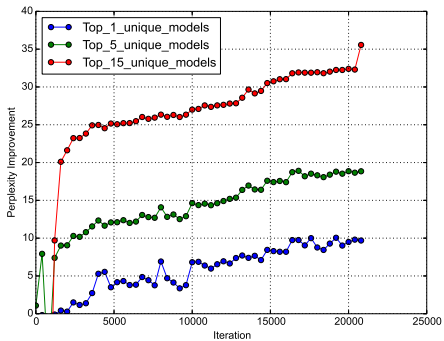}
\caption{Improvement of Neural Architecture Search over random search over time. We plot the difference between the average of the top $k$ models our controller finds vs. random search every 400 models run.}
\label{fig:learning_curve}
\end{center}
\end{figure}

\paragraph{Control Experiment 2 -- Comparison against Random Search:} Instead of policy gradient, one can use random search to find the best network. Although this baseline seems simple, it is often very hard to surpass~\citep{bergstra2012random}. We report the perplexity improvements using policy gradient against random search as training progresses in Figure~\ref{fig:learning_curve}. The results show that not only the best model using policy gradient is better than the best model using random search, but also the average of top models is also much better.

%An illustration of how this process works with a Base of 4 is shown in Figure~\ref{figure:RNN_Method}. 

%\subsection{Training Details}

\section{Conclusion}

In this paper we introduce Neural Architecture Search, an idea of using a recurrent neural network to compose neural network architectures. By using recurrent network as the controller, our method is flexible so that it can search variable-length architecture space. Our method has strong empirical performance on very challenging benchmarks and presents a new research direction for automatically finding good neural network architectures. The code for running the models found by the controller on CIFAR-10 and PTB will be released at https://github.com/tensorflow/models . Additionally, we have added the RNN cell found using our method under the name NASCell into TensorFlow, so others can easily use it.

\subsubsection*{Acknowledgments}
We thank Greg Corrado, Jeff Dean, David Ha, Lukasz Kaiser and the Google Brain team for their help with the project.

\bibliography{iclr2017_conference}

\begin{thebibliography}{66}
\providecommand{\natexlab}[1]{#1}
\providecommand{\url}[1]{\texttt{#1}}
\expandafter\ifx\csname urlstyle\endcsname\relax
  \providecommand{\doi}[1]{doi: #1}\else
  \providecommand{\doi}{doi: \begingroup \urlstyle{rm}\Url}\fi

\bibitem[Andreas et~al.(2016)Andreas, Rohrbach, Darrell, and
  Klein]{andreas2016learning}
Jacob Andreas, Marcus Rohrbach, Trevor Darrell, and Dan Klein.
\newblock Learning to compose neural networks for question answering.
\newblock In \emph{NAACL}, 2016.

\bibitem[Andrychowicz et~al.(2016)Andrychowicz, Denil, Gomez, Hoffman, Pfau,
  Schaul, and de~Freitas]{andrychowicz2016learning}
Marcin Andrychowicz, Misha Denil, Sergio Gomez, Matthew~W Hoffman, David Pfau,
  Tom Schaul, and Nando de~Freitas.
\newblock Learning to learn by gradient descent by gradient descent.
\newblock \emph{arXiv preprint arXiv:1606.04474}, 2016.

\bibitem[Bahdanau et~al.(2015)Bahdanau, Cho, and Bengio]{bahdanau2014neural}
Dzmitry Bahdanau, Kyunghyun Cho, and Yoshua Bengio.
\newblock Neural machine translation by jointly learning to align and
  translate.
\newblock In \emph{ICLR}, 2015.

\bibitem[Bengio et~al.(2003)Bengio, Ducharme, Vincent, and
  Jauvin]{bengio2003neural}
Yoshua Bengio, R{\'e}jean Ducharme, Pascal Vincent, and Christian Jauvin.
\newblock A neural probabilistic language model.
\newblock \emph{JMLR}, 2003.

\bibitem[Bergstra \& Bengio(2012)Bergstra and Bengio]{bergstra2012random}
James Bergstra and Yoshua Bengio.
\newblock Random search for hyper-parameter optimization.
\newblock \emph{JMLR}, 2012.

\bibitem[Bergstra et~al.(2011)Bergstra, Bardenet, Bengio, and
  K{\'e}gl]{bergstra2011algorithms}
James Bergstra, R{\'e}mi Bardenet, Yoshua Bengio, and Bal{\'a}zs K{\'e}gl.
\newblock Algorithms for hyper-parameter optimization.
\newblock In \emph{NIPS}, 2011.

\bibitem[Bergstra et~al.(2013)Bergstra, Yamins, and Cox]{bergstra2013making}
James Bergstra, Daniel Yamins, and David~D Cox.
\newblock Making a science of model search: Hyperparameter optimization in
  hundreds of dimensions for vision architectures.
\newblock \emph{ICML}, 2013.

\bibitem[Biermann(1978)]{biermann1978inference}
Alan~W. Biermann.
\newblock The inference of regular {LISP} programs from examples.
\newblock \emph{IEEE transactions on Systems, Man, and Cybernetics}, 1978.

\bibitem[Cheng et~al.(2014)Cheng, Kok, Pham, Chieu, and Chai]{Cheng2014}
Wei-Chen Cheng, Stanley Kok, Hoai~Vu Pham, Hai~Leong Chieu, and Kian Ming~Adam
  Chai.
\newblock Language modeling with sum-product networks.
\newblock In \emph{INTERSPEECH}, 2014.

\bibitem[Dalal \& Triggs(2005)Dalal and Triggs]{dalal2005histograms}
Navneet Dalal and Bill Triggs.
\newblock Histograms of oriented gradients for human detection.
\newblock In \emph{CVPR}, 2005.

\bibitem[Dean et~al.(2012)Dean, Corrado, Monga, Chen, Devin, Mao, Senior,
  Tucker, Yang, Le, et~al.]{dean2012large}
Jeffrey Dean, Greg Corrado, Rajat Monga, Kai Chen, Matthieu Devin, Mark Mao,
  Andrew Senior, Paul Tucker, Ke~Yang, Quoc~V. Le, et~al.
\newblock Large scale distributed deep networks.
\newblock In \emph{NIPS}, 2012.

\bibitem[Floreano et~al.(2008)Floreano, D{\"u}rr, and
  Mattiussi]{floreano2008neuroevolution}
Dario Floreano, Peter D{\"u}rr, and Claudio Mattiussi.
\newblock Neuroevolution: from architectures to learning.
\newblock \emph{Evolutionary Intelligence}, 2008.

\bibitem[Gal(2015)]{Gal2015}
Yarin Gal.
\newblock A theoretically grounded application of dropout in recurrent neural
  networks.
\newblock \emph{arXiv preprint arXiv:1512.05287}, 2015.

\bibitem[Ha et~al.(2016)Ha, Dai, and Le]{ha2016hypernetworks}
David Ha, Andrew Dai, and Quoc~V. Le.
\newblock Hypernetworks.
\newblock \emph{arXiv preprint arXiv:1609.09106}, 2016.

\bibitem[He et~al.(2016{\natexlab{a}})He, Zhang, Ren, and Sun]{he2015deep}
Kaiming He, Xiangyu Zhang, Shaoqing Ren, and Jian Sun.
\newblock Deep residual learning for image recognition.
\newblock In \emph{CVPR}, 2016{\natexlab{a}}.

\bibitem[He et~al.(2016{\natexlab{b}})He, Zhang, Ren, and
  Sun]{identity-mappings}
Kaiming He, Xiangyu Zhang, Shaoqing Ren, and Jian Sun.
\newblock Identity mappings in deep residual networks.
\newblock \emph{arXiv preprint arXiv:1603.05027}, 2016{\natexlab{b}}.

\bibitem[Hinton et~al.(2012)Hinton, Deng, Yu, Dahl, Mohamed, Jaitly, Senior,
  Vanhoucke, Nguyen, Sainath, et~al.]{hinton2012deep}
Geoffrey Hinton, Li~Deng, Dong Yu, George~E. Dahl, Abdel-rahman Mohamed,
  Navdeep Jaitly, Andrew Senior, Vincent Vanhoucke, Patrick Nguyen, Tara~N.
  Sainath, et~al.
\newblock Deep neural networks for acoustic modeling in speech recognition: The
  shared views of four research groups.
\newblock \emph{IEEE Signal Processing Magazine}, 2012.

\bibitem[Hochreiter \& Schmidhuber(1997)Hochreiter and Schmidhuber]{lstm}
Sepp Hochreiter and Juergen Schmidhuber.
\newblock Long short-term memory.
\newblock \emph{Neural Computation}, 1997.

\bibitem[Huang et~al.(2016{\natexlab{a}})Huang, Liu, and
  Weinberger]{Huang2016Densely}
Gao Huang, Zhuang Liu, and Kilian~Q. Weinberger.
\newblock Densely connected convolutional networks.
\newblock \emph{arXiv preprint arXiv:1608.06993}, 2016{\natexlab{a}}.

\bibitem[Huang et~al.(2016{\natexlab{b}})Huang, Liu, Weinberger, and van~der
  Maaten]{Huang2016Densely2}
Gao Huang, Zhuang Liu, Kilian~Q. Weinberger, and Laurens van~der Maaten.
\newblock Densely connected convolutional networks.
\newblock \emph{arXiv preprint arXiv:1608.06993}, 2016{\natexlab{b}}.

\bibitem[Huang et~al.(2016{\natexlab{c}})Huang, Sun, Liu, Sedra, and
  Weinberger]{stochastic}
Gao Huang, Yu~Sun, Zhuang Liu, Daniel Sedra, and Kilian Weinberger.
\newblock Deep networks with stochastic depth.
\newblock \emph{arXiv preprint arXiv:1603.09382}, 2016{\natexlab{c}}.

\bibitem[Inan et~al.(2016)Inan, Khosravi, and Socher]{SocherEmbedding}
Hakan Inan, Khashayar Khosravi, and Richard Socher.
\newblock Tying word vectors and word classifiers: A loss framework for
  language modeling.
\newblock \emph{arXiv preprint arXiv:1611.01462}, 2016.

\bibitem[Ioffe \& Szegedy(2015)Ioffe and Szegedy]{BatchNorm}
Sergey Ioffe and Christian Szegedy.
\newblock Batch normalization: Accelerating deep network training by reducing
  internal covariate shift.
\newblock In \emph{ICML}, 2015.

\bibitem[Jarrett et~al.(2009)Jarrett, Kavukcuoglu, Lecun,
  et~al.]{jarrett2009best}
Kevin Jarrett, Koray Kavukcuoglu, Yann Lecun, et~al.
\newblock What is the best multi-stage architecture for object recognition?
\newblock In \emph{ICCV}, 2009.

\bibitem[Jozefowicz et~al.(2015)Jozefowicz, Zaremba, and
  Sutskever]{jozefowicz2015empirical}
Rafal Jozefowicz, Wojciech Zaremba, and Ilya Sutskever.
\newblock An empirical exploration of recurrent network architectures.
\newblock In \emph{ICML}, 2015.

\bibitem[Kim et~al.(2015)Kim, Jernite, Sontag, and Rush]{Kim2016}
Yoon Kim, Yacine Jernite, David Sontag, and Alexander~M. Rush.
\newblock Character-aware neural language models.
\newblock \emph{arXiv preprint arXiv:1508.06615}, 2015.

\bibitem[Kingma \& Ba(2015)Kingma and Ba]{ADAM}
Diederik~P. Kingma and Jimmy Ba.
\newblock Adam: {A} method for stochastic optimization.
\newblock In \emph{ICLR}, 2015.

\bibitem[Krizhevsky et~al.(2012)Krizhevsky, Sutskever, and
  Hinton]{krizhevsky2012imagenet}
Alex Krizhevsky, Ilya Sutskever, and Geoffrey~E. Hinton.
\newblock Imagenet classification with deep convolutional neural networks.
\newblock In \emph{NIPS}, 2012.

\bibitem[Lake et~al.(2015)Lake, Salakhutdinov, and Tenenbaum]{lake2015human}
Brenden~M. Lake, Ruslan Salakhutdinov, and Joshua~B. Tenenbaum.
\newblock Human-level concept learning through probabilistic program induction.
\newblock \emph{Science}, 2015.

\bibitem[Larsson et~al.(2016)Larsson, Maire, and Shakhnarovich]{fractalnet}
Gustav Larsson, Michael Maire, and Gregory Shakhnarovich.
\newblock Fractalnet: Ultra-deep neural networks without residuals.
\newblock \emph{arXiv preprint arXiv:1605.07648}, 2016.

\bibitem[LeCun et~al.(1998)LeCun, Bottou, Bengio, and
  Haffner]{lecun1998gradient}
Yann LeCun, L{\'e}on Bottou, Yoshua Bengio, and Patrick Haffner.
\newblock Gradient-based learning applied to document recognition.
\newblock \emph{Proceedings of the IEEE}, 1998.

\bibitem[Lee et~al.(2015)Lee, Xie, Gallagher, Zhang, and Tu]{dsn}
Chen-Yu Lee, Saining Xie, Patrick Gallagher, Zhengyou Zhang, and Zhuowen Tu.
\newblock Deeply-supervised nets.
\newblock In \emph{AISTATS}, 2015.

\bibitem[Li \& Malik(2016)Li and Malik]{li2016learning}
Ke~Li and Jitendra Malik.
\newblock Learning to optimize.
\newblock \emph{arXiv preprint arXiv:1606.01885}, 2016.

\bibitem[Liang et~al.(2010)Liang, Jordan, and Klein]{liang2010learning}
Percy Liang, Michael~I. Jordan, and Dan Klein.
\newblock Learning programs: A hierarchical {B}ayesian approach.
\newblock In \emph{ICML}, 2010.

\bibitem[Lin et~al.(2013)Lin, Chen, and Yan]{netinnet}
Min Lin, Qiang Chen, and Shuicheng Yan.
\newblock Network in network.
\newblock In \emph{ICLR}, 2013.

\bibitem[Lowe(1999)]{lowe1999object}
David~G. Lowe.
\newblock Object recognition from local scale-invariant features.
\newblock In \emph{CVPR}, 1999.

\bibitem[Mendoza et~al.(2016)Mendoza, Klein, Feurer, Springenberg, and
  Hutter]{mendoza2016towards}
Hector Mendoza, Aaron Klein, Matthias Feurer, Jost~Tobias Springenberg, and
  Frank Hutter.
\newblock Towards automatically-tuned neural networks.
\newblock In \emph{Proceedings of the 2016 Workshop on Automatic Machine
  Learning}, pp.\  58--65, 2016.

\bibitem[Merity et~al.(2016)Merity, Xiong, Bradbury, and Socher]{Socher2016}
Stephen Merity, Caiming Xiong, James Bradbury, and Richard Socher.
\newblock Pointer sentinel mixture models.
\newblock \emph{arXiv preprint arXiv:1609.07843}, 2016.

\bibitem[Mikolov \& Zweig(2012)Mikolov and Zweig]{Mikolov2012}
Tomas Mikolov and Geoffrey Zweig.
\newblock Context dependent recurrent neural network language model.
\newblock In \emph{SLT}, pp.\  234--239, 2012.

\bibitem[Mnih \& Hinton(2007)Mnih and Hinton]{mnih2007three}
Andriy Mnih and Geoffrey Hinton.
\newblock Three new graphical models for statistical language modelling.
\newblock In \emph{ICML}, 2007.

\bibitem[Nair \& Hinton(2010)Nair and Hinton]{nair2010rectified}
Vinod Nair and Geoffrey~E. Hinton.
\newblock Rectified linear units improve restricted {B}oltzmann machines.
\newblock In \emph{ICML}, 2010.

\bibitem[Neelakantan et~al.(2015)Neelakantan, Le, and
  Sutskever]{neelakantan2015neural}
Arvind Neelakantan, Quoc~V. Le, and Ilya Sutskever.
\newblock Neural programmer: Inducing latent programs with gradient descent.
\newblock In \emph{ICLR}, 2015.

\bibitem[Pascanu et~al.(2013)Pascanu, Gulcehre, Cho, and Bengio]{Pascanu2013a}
Razvan Pascanu, Caglar Gulcehre, Kyunghyun Cho, and Yoshua Bengio.
\newblock How to construct deep recurrent neural networks.
\newblock \emph{arXiv preprint arXiv:1312.6026}, 2013.

\bibitem[Press \& Wolf(2016)Press and Wolf]{shareEmbedding}
Ofir Press and Lior Wolf.
\newblock Using the output embedding to improve language models.
\newblock \emph{arXiv preprint arXiv:1608.05859}, 2016.

\bibitem[Ranzato et~al.(2015)Ranzato, Chopra, Auli, and
  Zaremba]{ranzato2015sequence}
Marc'Aurelio Ranzato, Sumit Chopra, Michael Auli, and Wojciech Zaremba.
\newblock Sequence level training with recurrent neural networks.
\newblock \emph{arXiv preprint arXiv:1511.06732}, 2015.

\bibitem[Reed \& de~Freitas(2015)Reed and de~Freitas]{reed2015neural}
Scott Reed and Nando de~Freitas.
\newblock Neural programmer-interpreters.
\newblock In \emph{ICLR}, 2015.

\bibitem[Saxena \& Verbeek(2016)Saxena and Verbeek]{saxena2016convolutional}
Shreyas Saxena and Jakob Verbeek.
\newblock Convolutional neural fabrics.
\newblock In \emph{NIPS}, 2016.

\bibitem[Shen et~al.(2016)Shen, Cheng, He, He, Wu, Sun, and Liu]{ShenCHHWSL15}
Shiqi Shen, Yong Cheng, Zhongjun He, Wei He, Hua Wu, Maosong Sun, and Yang Liu.
\newblock Minimum risk training for neural machine translation.
\newblock In \emph{ACL}, 2016.

\bibitem[Simonyan \& Zisserman(2014)Simonyan and Zisserman]{simonyan2014very}
Karen Simonyan and Andrew Zisserman.
\newblock Very deep convolutional networks for large-scale image recognition.
\newblock \emph{arXiv preprint arXiv:1409.1556}, 2014.

\bibitem[Snoek et~al.(2012)Snoek, Larochelle, and Adams]{snoek2012practical}
Jasper Snoek, Hugo Larochelle, and Ryan~P. Adams.
\newblock Practical {B}ayesian optimization of machine learning algorithms.
\newblock In \emph{NIPS}, 2012.

\bibitem[Snoek et~al.(2015)Snoek, Rippel, Swersky, Kiros, Satish, Sundaram,
  Patwary, Ali, Adams, et~al.]{snoek2015scalable}
Jasper Snoek, Oren Rippel, Kevin Swersky, Ryan Kiros, Nadathur Satish,
  Narayanan Sundaram, Mostofa Patwary, Mostofa Ali, Ryan~P. Adams, et~al.
\newblock Scalable bayesian optimization using deep neural networks.
\newblock In \emph{ICML}, 2015.

\bibitem[Springenberg et~al.(2014)Springenberg, Dosovitskiy, Brox, and
  Riedmiller]{allcnn}
Jost~Tobias Springenberg, Alexey Dosovitskiy, Thomas Brox, and Martin
  Riedmiller.
\newblock Striving for simplicity: The all convolutional net.
\newblock \emph{arXiv preprint arXiv:1412.6806}, 2014.

\bibitem[Srivastava et~al.(2015)Srivastava, Greff, and Schmidhuber]{highway}
Rupesh~Kumar Srivastava, Klaus Greff, and J{\"u}rgen Schmidhuber.
\newblock Highway networks.
\newblock \emph{arXiv preprint arXiv:1505.00387}, 2015.

\bibitem[Stanley et~al.(2009)Stanley, D'Ambrosio, and
  Gauci]{stanley2009hypercube}
Kenneth~O. Stanley, David~B. D'Ambrosio, and Jason Gauci.
\newblock A hypercube-based encoding for evolving large-scale neural networks.
\newblock \emph{Artificial {L}ife}, 2009.

\bibitem[Summers(1977)]{summers1977methodology}
Phillip~D. Summers.
\newblock A methodology for {LISP} program construction from examples.
\newblock \emph{Journal of the ACM}, 1977.

\bibitem[Sutskever et~al.(2013)Sutskever, Martens, Dahl, and
  Hinton]{icml2013_sutskever13}
Ilya Sutskever, James Martens, George Dahl, and Geoffrey Hinton.
\newblock On the importance of initialization and momentum in deep learning.
\newblock In \emph{ICML}, 2013.

\bibitem[Sutskever et~al.(2014)Sutskever, Vinyals, and
  Le]{sutskever2014sequence}
Ilya Sutskever, Oriol Vinyals, and Quoc~V. Le.
\newblock Sequence to sequence learning with neural networks.
\newblock In \emph{NIPS}, 2014.

\bibitem[Szegedy et~al.(2015)Szegedy, Liu, Jia, Sermanet, Reed, Anguelov,
  Erhan, Vanhoucke, and Rabinovich]{szegedy2015going}
Christian Szegedy, Wei Liu, Yangqing Jia, Pierre Sermanet, Scott Reed, Dragomir
  Anguelov, Dumitru Erhan, Vincent Vanhoucke, and Andrew Rabinovich.
\newblock Going deeper with convolutions.
\newblock In \emph{CVPR}, 2015.

\bibitem[Thrun \& Pratt(2012)Thrun and Pratt]{thrun2012learning}
Sebastian Thrun and Lorien Pratt.
\newblock \emph{Learning to learn}.
\newblock Springer Science \& Business Media, 2012.

\bibitem[Vinyals et~al.(2015)Vinyals, Fortunato, and
  Jaitly]{vinyals2015pointer}
Oriol Vinyals, Meire Fortunato, and Navdeep Jaitly.
\newblock Pointer networks.
\newblock In \emph{NIPS}, 2015.

\bibitem[Wierstra et~al.(2005)Wierstra, Gomez, and
  Schmidhuber]{wierstra2005modeling}
Daan Wierstra, Faustino~J Gomez, and J{\"u}rgen Schmidhuber.
\newblock Modeling systems with internal state using evolino.
\newblock In \emph{GECCO}, 2005.

\bibitem[Williams(1992)]{Williams92simplestatistical}
Ronald~J. Williams.
\newblock Simple statistical gradient-following algorithms for connectionist
  reinforcement learning.
\newblock In \emph{Machine Learning}, 1992.

\bibitem[Wu et~al.(2016)Wu, Schuster, Chen, Le, Norouzi, et~al.]{wu2016google}
Yonghui Wu, Mike Schuster, Zhifeng Chen, Quoc~V. Le, Mohammad Norouzi, et~al.
\newblock Google's neural machine translation system: Bridging the gap between
  human and machine translation.
\newblock \emph{arXiv preprint arXiv:1609.08144}, 2016.

\bibitem[Zagoruyko \& Komodakis(2016)Zagoruyko and Komodakis]{wide}
Sergey Zagoruyko and Nikos Komodakis.
\newblock Wide residual networks.
\newblock In \emph{BMVC}, 2016.

\bibitem[Zaremba et~al.(2014)Zaremba, Sutskever, and Vinyals]{ZarembaReg}
Wojciech Zaremba, Ilya Sutskever, and Oriol Vinyals.
\newblock Recurrent neural network regularization.
\newblock \emph{arXiv preprint arXiv:1409.2329}, 2014.

\bibitem[Zilly et~al.(2016)Zilly, Srivastava, Koutn{\'\i}k, and
  Schmidhuber]{Zilly2016}
Julian~Georg Zilly, Rupesh~Kumar Srivastava, Jan Koutn{\'\i}k, and J{\"u}rgen
  Schmidhuber.
\newblock Recurrent highway networks.
\newblock \emph{arXiv preprint arXiv:1607.03474}, 2016.

\end{thebibliography}
\bibliographystyle{iclr2017_conference}

\newpage
\appendix
\section{Appendix}
\label{sec:appendix}

\begin{figure}[h!]
\begin{center}
\centerline{\includegraphics[scale=0.6]{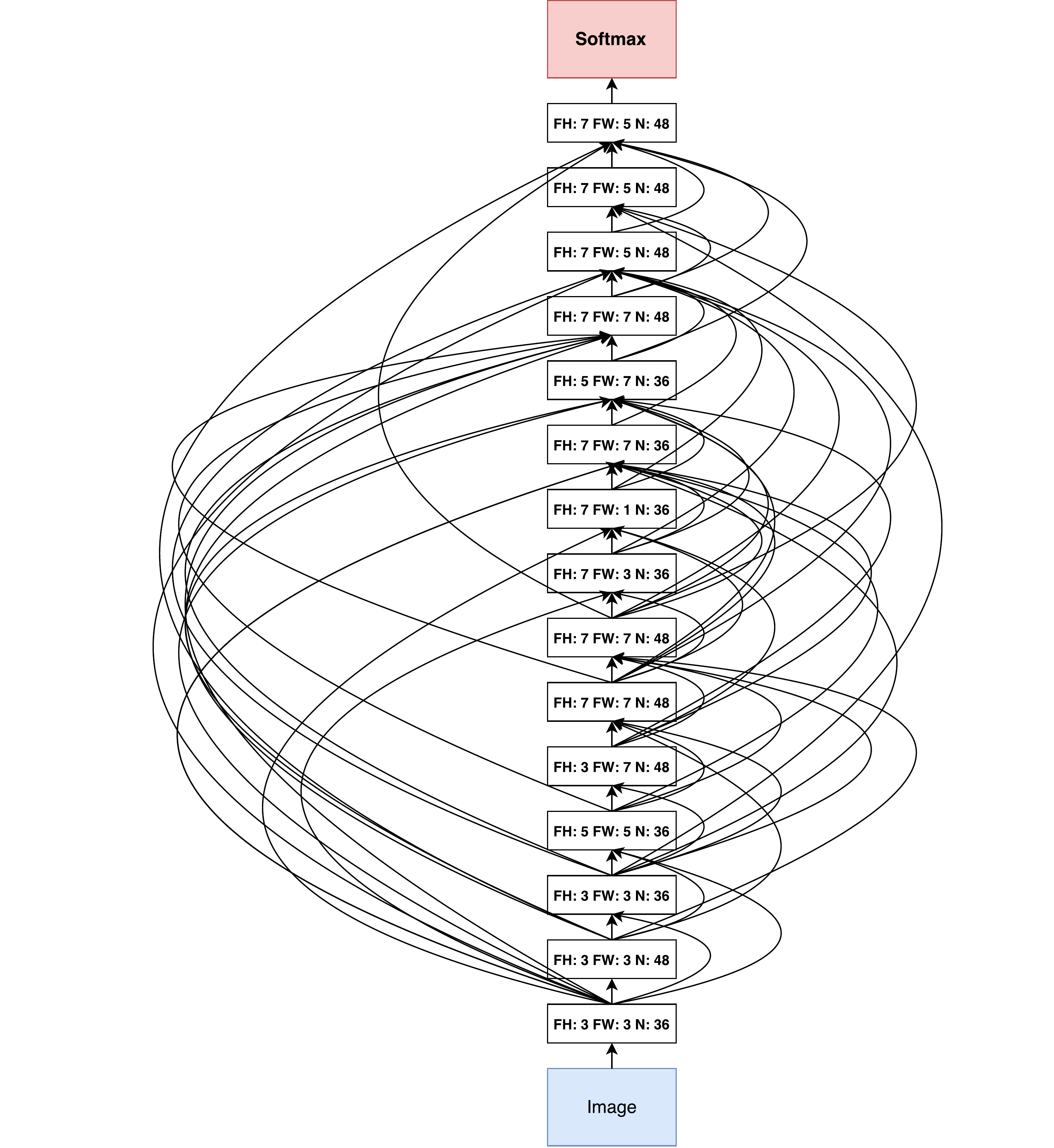}}
\caption{Convolutional architecture discovered by our method, when the
  search space does not have strides or pooling layers. FH is filter
  height, FW is filter width and N is number of filters. Note that the
  skip connections are not residual connections. If one layer has many
  input layers then all input layers are concatenated in the depth
  dimension.}
\label{figure:strange_net}
\end{center}
\end{figure}

\begin{figure}
\begin{center}
  \includegraphics[width=1.0\textwidth]{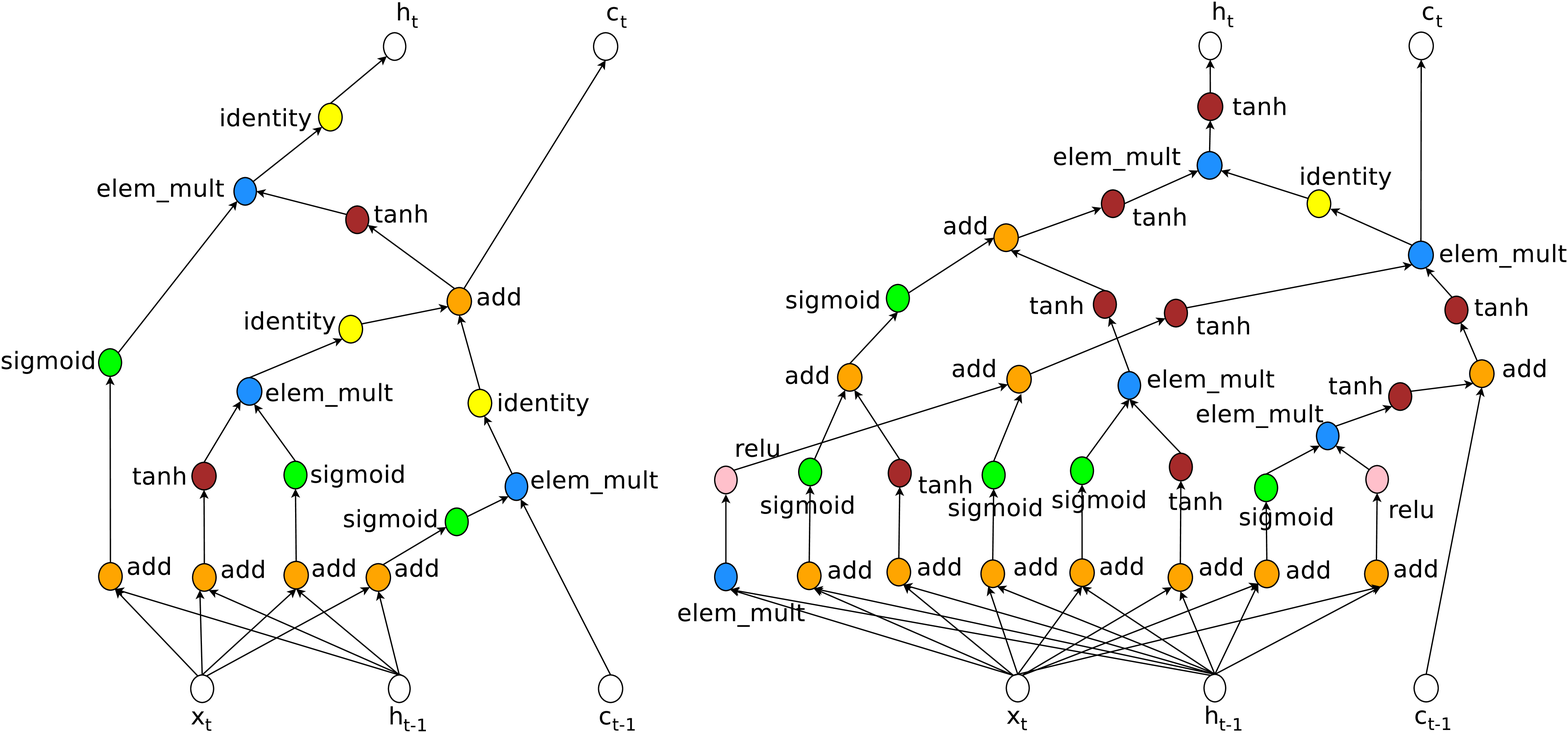}
  \includegraphics[width=0.5\textwidth]{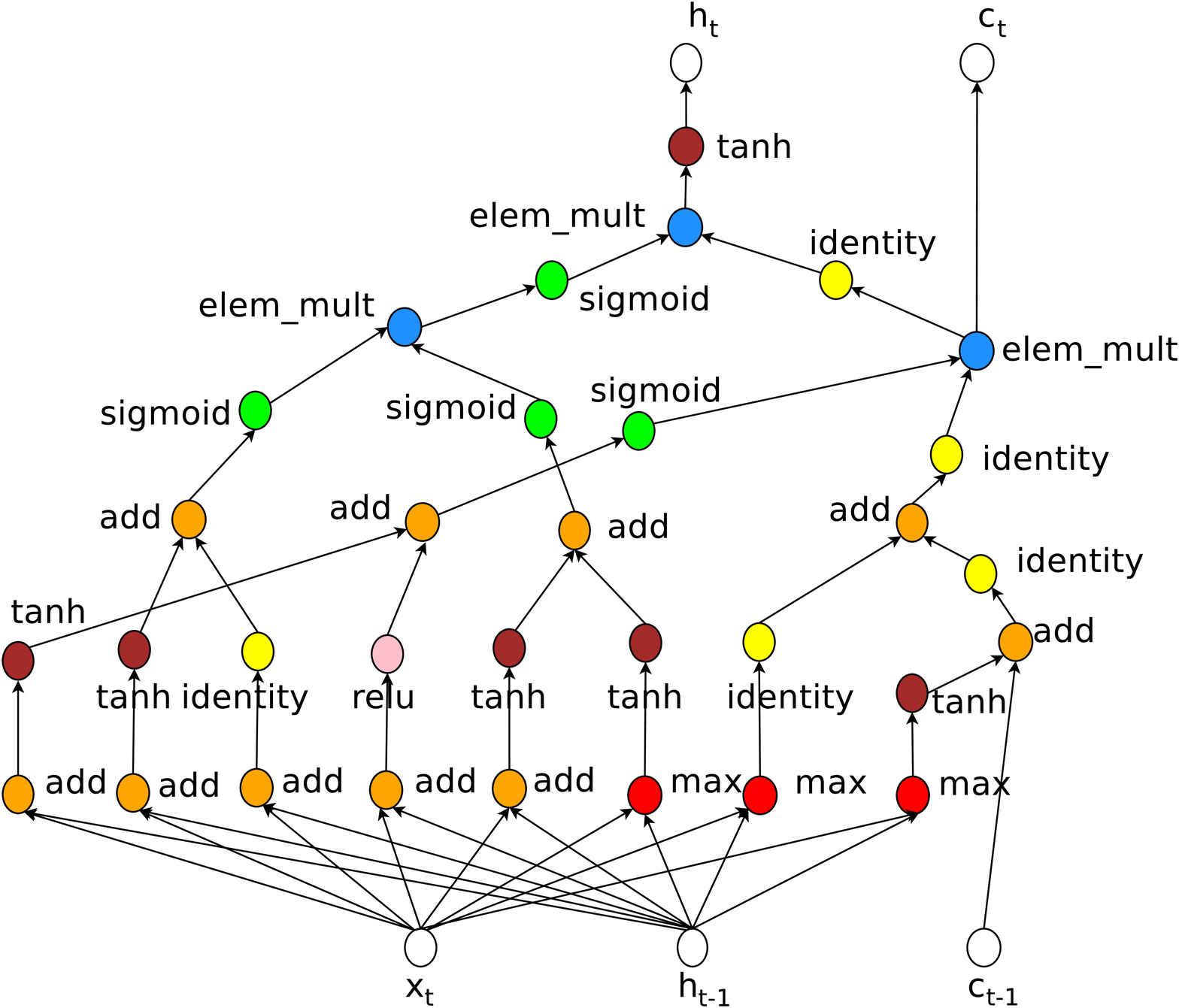}
\caption{A comparison of the original LSTM cell vs. two good cells our
  model found. Top left: LSTM cell. Top right: Cell found by our model
  when the search space does not include $max$ and $sin$. Bottom: Cell
  found by our model when the search space includes $max$ and $sin$
  (the controller did not choose to use the $sin$ function).}
\label{fig:strange_cell}
\end{center}
\end{figure}

\end{document}